%% file: ms.tex
\documentclass[10pt,twocolumn,letterpaper]{article}
\pdfoutput=1

\usepackage{times}
\usepackage{epsfig}
\usepackage{graphicx}
\usepackage{amsmath}
\usepackage{amssymb}
\usepackage{authblk}

\input{our_preamble.tex}

\usepackage[pagebackref=true,breaklinks=true,colorlinks,citecolor=blue,bookmarks=true]{hyperref}

\begin{document}

\title{Minority Reports Defense:\\
Defending Against Adversarial Patches}

\renewcommand\Authsep{ \hspace{1.4em}}
\renewcommand\Authands{ \hspace{1.4em}}
\author[1]{Michael McCoyd}
\author[2]{Won Park\thanks{Work done while Park was at the University of California, Berkeley.}}
\author[1]{Steven Chen}
\author[1]{Neil Shah}
\author[1]{Ryan Roggenkemper}
\author[1]{Minjune Hwang}
\author[1]{Jason Xinyu Liu}
\author[1]{David Wagner}
\affil[1]{University of California, Berkeley}
\affil[2]{University of Michigan}

\date{}
\maketitle

\begin{abstract}
\input{abstract.tex}
\end{abstract}


\input{Intro}
\input{models}
\input{attack}

\input{defense}
\input{security}
\input{higher_res}

\input{defense_experiments} 
\input{accuracy_effects}

\input{related}


\input{conclusion}

{\small
\bibliographystyle{halpha}
\bibliography{../AdversarialExamples.bib}
}

\end{document}

%% file: our_preamble.tex
\usepackage{graphicx} 
\usepackage{amsmath}
\usepackage{amssymb} 

\usepackage{geometry}
\usepackage[protrusion=true,  expansion, tracking=true, spacing=true]{microtype} 
\usepackage{array} 

\usepackage{caption}
\usepackage{subcaption} 

\usepackage{url}

\usepackage{rotating}    
\usepackage{color}
\usepackage{etoolbox}    

\usepackage{booktabs} 
\usepackage{comment}

\usepackage{xspace}    

\usepackage[linesnumbered,ruled, vlined, algo2e]{algorithm2e}

\usepackage{paralist}

\usepackage{fancyvrb}  

\usepackage{refstyle} 

\newtoggle{comments}     
\settoggle{comments}{true}
\iftoggle{comments}{
   \newrobustcmd{\full}[1]{{\color{gray}#1}}
   \newrobustcmd*{\tocite}[1]{{\footnotesize\color{blue}[#1]}}
   \newrobustcmd{\todo}[1]{{\footnotesize\color{blue}[#1]}}
   \newrobustcmd{\mjm}[1]{{\scriptsize\color{blue}[mjm: #1]}}
   \newrobustcmd{\add}[1]{{\footnotesize\color{green}[TO ADD: #1]}}
   \newrobustcmd{\daw}[1]{{\footnotesize\color{magenta}[daw: #1]}}
}{
   \newrobustcmd{\full}[1]{}
   \newrobustcmd*{\tocite}[1]{}
   \newrobustcmd{\todo}[1]{}
   \newrobustcmd{\mjm}[1]{}
   \newrobustcmd{\daw}[1]{}
}

\makeatletter
\renewcommand\paragraph{%
 \@startsection {paragraph}{4}{\z@}{.4ex plus 0.1ex minus 0.1ex}{-1em}%
{\normalfont \normalsize \bfseries }}%
\makeatother

%% file: abstract.tex
Deep learning image classification is vulnerable to adversarial attack,
  even if the attacker changes just a small patch of the image.
We propose a defense against patch attacks based on
  partially occluding the image around each candidate patch location,
  so that a few occlusions each completely hide the patch.
We demonstrate on CIFAR-10, Fashion MNIST, and MNIST that our defense
provides certified security against patch attacks of a certain size.

%% file: Intro.tex


\section{Introduction}
An attacker with knowledge of a neural network model can construct, 
from any normal image $x$, an adversarial example $x^\star$ 
 that looks to humans like $x$
 but that the model classifies differently from the normal image
 \cite{Sz-2013-Int, Good-2014-exp, Huang11-adversarialmachine, carlini2017towards}.

Recently, researchers have proposed the adversarial patch attack \cite{brown2017-advpatch, karmon2017-LaVAN},
where the attacker changes just a limited rectangular region of the image,
for example by placing a sticker over a road sign or other object.
Others have expanded on the vulnerability to this type of attack
\cite{eykholt2017robust, Thys-patch, xu2019adversarial}.
In this paper, we propose a defense against this attack.

The idea of our defense is to occlude part of the image and then
classify the occluded image.
First, we train a classifier that properly classifies occluded images.
Then, if we knew the location of the adversarial patch,
we could occlude that region of
the image (e.g., overwriting it with a uniform grey rectangle)
and apply the classifier to the occluded image.
This would defend against patch attacks, as the attacker's contribution
is completely overwritten and the input to the classifier (the occluded
image) cannot be affected by the attacker in any way.

\input{figs/fig_intro}

In practice, we do not know the location of the adversarial patch,
so a more sophisticated defense is needed.
Our approach works by 
  occluding an area larger than the maximum patch size
  and striding the occlude area across the image,
  making an occluded prediction at each stride.
We then analyze the classifier's predictions on these occluded images.
If the occlusion region is sufficiently larger than the adversarial patch,
several of the occluded images will completely obscure the adversarial
patch and thus the classifier's prediction on those images will be
unaffected by the adversary and should match the correct label.
Thus, we expect the correct label to appear multiple times among the
predictions from occluded images.
We show how to use this redundancy to detect adversarial patch attacks.
We call our scheme the minority reports defense
because no matter where the patch is located,
there will always be a minority of predictions that cannot be
influenced by the attacker and vote for the correct label.

\Figref{fig:into:attack} illustrates our defense.
We take the input image (\Figref{fig:into:attack_full})
  and construct
  a grid of partially occluded images (\Figref{fig:into:attack_obscure})
  with occlusions at different locations,
  chosen so that any attack will be occluded in a cluster of several
  predictions.
We then apply the classifier to each occluded image to obtain a grid of predictions.
When under attack, we can expect most predictions to differ from the
true label, but there will always be a cluster of locations where the
adversarial patch is fully obscured and thus the labels are all expected
to agree with the true label;
in \Figref{fig:into:attack}, the 3rd and 4th images in the 4th row obscure
the adversarial patch and thus vote for the true label.
Our defense analyzes the grid of predicted labels to detect this pattern.
If there is a cluster of predictions that all match each other but are in the minority for the prediction grid overall, then this suggests an attack.
\Figref{fig:intro:grid_examples} visualizes the prediction grid for
a benign image (on the left) and a malicious image containing an 
undefended adversarial patch (on the right).%
\footnote{For the eagle eyed reader, our illustrations are still for a $6 \times 6$ patch.}

We evaluate our scheme on the 
  CIFAR-10 \cite{krizhevsky2009learning}, 
  Fashion MNIST \cite{xiao_etal_2017_fashion}, 
  and MNIST \cite{lecun-98-gradi-mnist} 
  datasets with a stride of one.
We show that our defense does not harm accuracy much.
We also evaluate its security against adaptive attacks.
In particular, we show how to bound the success of any possible
attack on a given image, and using this we are able to demonstrate
certified security for a large fraction of images.
In particular, we are able to prove a security theorem: for
a large fraction of images in the validation set, we can prove that
no patch attack will succeed, no matter where the patch is placed
or how the patch is modified, so long as the size of the patch is
limited.

Our contributions are:
\begin{compactitem}
\item We quantify the vulnerability of undefended networks for
  Fashion MNIST and MNIST against patch attacks with patches of
  different sizes (\secref*{sec:feasibility}).
\item We propose a novel method for detecting patch attacks, 
  based on differently occluded views of the input image (\secref*{sec:defense}).
\item We provide a worst-case analysis of security against
  adaptive attacks for CIFAR-10, Fashion MNIST, and MNIST (\secref*{sec:security, sec:experiments}).
\end{compactitem}



%% file: figs/fig_intro.tex
\begin{figure}[tbp]\begin{center}

\begin{subfigure}[t]{\linewidth}
  \centering
\includegraphics[width=.19\linewidth]{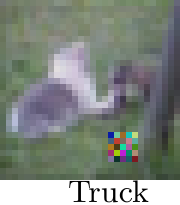}
  \subcaption{An attack image: a cat with a malicious $5 \times 5$ sticker that causes a standard model to classify it as a truck.}
  \figlabel{fig:into:attack_full}
\end{subfigure}

\begin{subfigure}[t]{\linewidth}
  \centering
\includegraphics[width=.83\linewidth]{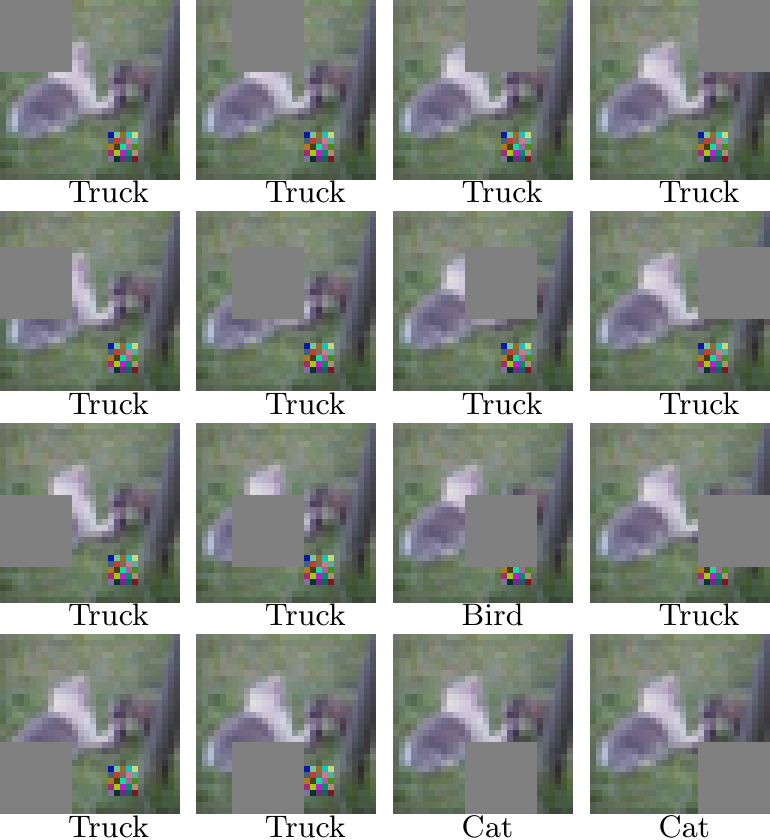}
  \subcaption{
  We occlude part of the image with a grey square, then classify these
  occluded images.
  Here the 3rd and 4th predictions in the 4th row will be unaffected by this attack.
  Our actual defense ensures that any attack will be fully occluded by a $3 \times 3$ grid of predictions,
  instead of the $1 \times 2$ grid shown here.
  }
   \figlabel{fig:into:attack_obscure}
\end{subfigure}

\caption{
Our scheme works by occluding different portions of the image and analyzing the predictions made by the classifier on these occluded images.
}
\figlabel{fig:into:attack}
\end{center}\end{figure}

\begin{figure}[tbp]\begin{center}
\begin{subfigure}[t]{0.4\linewidth}
  \centering
  \includegraphics[width=\linewidth]{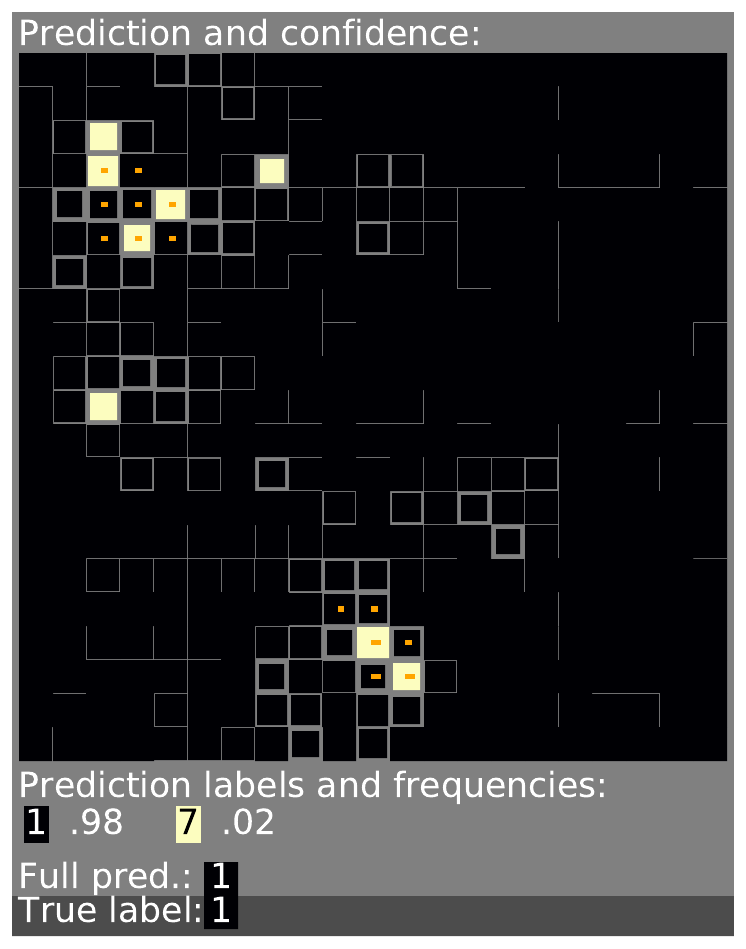}  
  \subcaption{Benign}
  \figlabel{fig:intro_eg_benign}
\end{subfigure}
\begin{subfigure}[t]{0.4\linewidth}
  \centering
  \includegraphics[width=\linewidth]{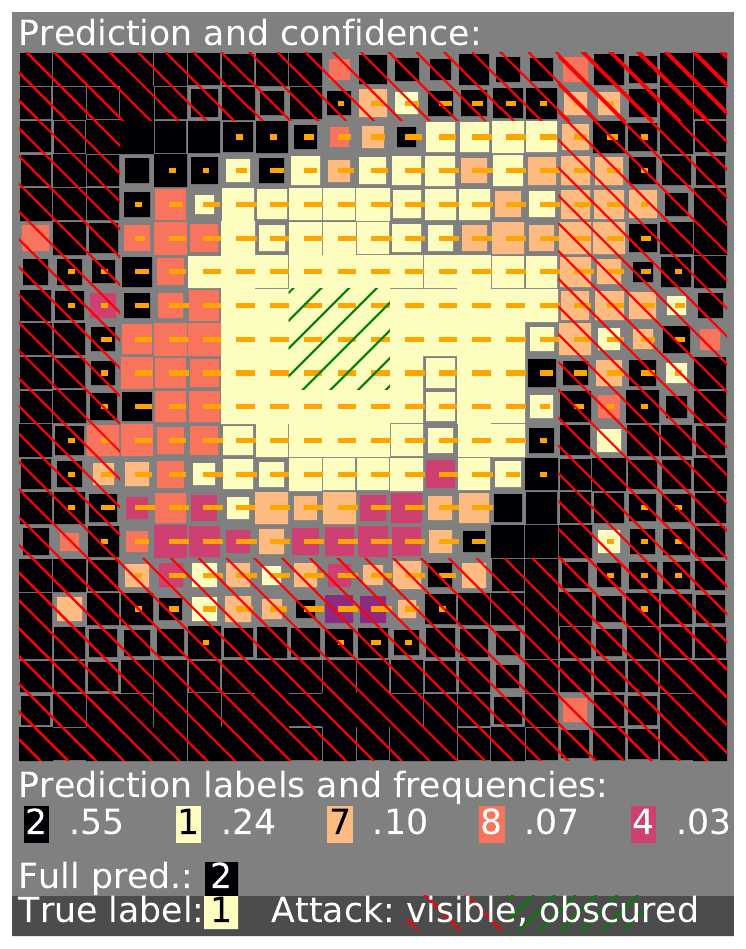}  
  \subcaption{Attack}
  \figlabel{fig:intro_eg_attack}
\end{subfigure}
\caption{
Prediction grids for a benign image (left) and an undefended attack image (right).
Each cell in the grid is colored based on the prediction made by
the classifier when fed an image obscured at that position in the grid.
A cluster of identical minority predictions, as seen in the right image,
suggests an attack.
In the attack image on the right, green hashes mark the nine predictions
where the adversarial patch was fully occluded.
}
\figlabel{fig:intro:grid_examples}
\end{center}\end{figure}

%% file: models.tex


\section{Data and Inner Model Training}\seclabel{sec:data_model}

Our defense sends partially occluded images to an inner model, 
  which returns a normal logit prediction for the dataset classes.
For this inner model, we use a standard convolutional architecture,
trained with data augmentation and random 90/10 train/validation splits:
for CIFAR-10, we use SimpNet's 600K parameter version
     \cite{HasanPour_etal_SimpNet_2018}
trained for 700 epochs, though we do not yet reproduce all details of their training;
for Fashion MNIST, a VGG-16 model \cite{VGG}
trained for 50 epochs;
for MNIST, the Deotte model \cite{Deotte}
 ([32C3-32C3-32C5S2] - [64C3-64C3-64C5S2] - 128),
with 40\% dropout and batch normalization
and 45 epochs.

As our defense will partially occlude the image, we train
these inner models with occluded images.
Each time an image is presented in training,
  a randomly placed $n \times n$ square is occluded
  and the model receives the occluded image.
This is similar to cutout from Devries et al. \cite{Devries_cutout_2107},
  who used occlusion as a regularizer.
We differ in that we also provide the model an additional input
  containing a sparsity mask that indicates which pixels are occluded.
For instance, the input to the CIFAR-10 model is
  an image, with dimensions $32 \times 32 \times 3$, and
  a mask, with dimensions $32 \times 32 \times 1$.
In the mask, a 0 indicates an occluded position and a 1
indicates a non-occluded position.
To better handle the missing pixels, convolutions in the inner model architecture
  are replaced with sparsity
  invariant convolutions \cite{uhrig-etal2017sparsity}.
If the mask indicates no occlusions,
  the sparsity invariant convolutions behave as normal convolutions;
  but, when occlusions are indicated, the occluded pixels are handled better.
We report inner model accuracies in \secref*{sec:accuracy_effects}.



%% file: attack.tex

\section{Patch Attack}\seclabel{sec:attack}

Patch attacks \cite{brown2017-advpatch} work by replacing a small
  part of the image with something of the attacker's choosing,
  e.g., by placing a small sticker on an object or road sign.
\Figref{fig:into:attack_full} shows a patch attack.
Patch attacks represent a practical method of executing an attack in the physical world.
It is not uncommon to see stickers on road signs in the real world,
  without preventing humans from understanding the signs nor
  prompting the immediate removal of the patch.
We see patch attacks partly as a practical concern,
  and partly as a stepping stone toward defending against full image attacks.

\subsection{Attack model}
We assume the attacker, with complete knowledge of the model,
   may select a square area of limited
  size anywhere within the digital image and arbitrarily modify
  all pixels within that square to any values in the pixel range.
For simplicity, we restrict the attacker to a square patch.
Our approach can handle other shapes as well so long as they are known
  in advance.

\subsection{Patch sizes}\seclabel{sec:feasibility}
We first studied how large a patch is needed to successfully attack our models.
We test multiple patch sizes and measure the attacker's
success rate for each patch size.
  
\paragraph{Setup}
We conduct a targeted attack against our Fashion MNIST and MNIST
  models from \Secref*{sec:data_model}.
We attack the first 300 validation images for Fashion MNIST and
  the first 100 validation images for MNIST, and
  report the fraction of images for which we are able to successfully
  mount a patch attack.
For each image, we select a target label by choosing randomly among the
  classes that are least likely, according to the softmax outputs
  of the classifier (namely, we find the least likely class, identify
  all classes whose confidence is within 0.1\% of the least likely,
  and select the target class uniformly at random among this set).
That target is used for all attacks on that image.
For each base image and its chosen target class,
 we enumerate all possible patch positions and try 
 at each position to find an attack patch at that position.
 
\paragraph{Attack algorithm}
To generate patch attacks, we iterate over all possible locations
  for the patch, and use a projected gradient descent (PGD) attack for each
  location.
We consider the attack a success if we find any location where we can
  place a patch that changes the model's prediction to the target label.
The resulting adversarial patch is specific to one particular image
  and one particular location.

The standard PGD attack uses a constant step size, but we found it was
  more effective to use a schedule that varies the step size among iterations.
In our experiments, a cyclic learning rate was more effective than a constant
  step size or a exponential decay rate, so we used it in all
  experiments.
We used a cyclic learning rate with 10 steps per cycle,
  with step sizes from 0.002 to 0.3, for a maximum of 150 steps.
We stopped early at the end of a cycle if the attack achieved
  confidence 0.6 or higher for the target class, or if the confidence
  had not improved by at least 0.002 in the last 20 steps from the best
  so far.
For each image, we attacked in parallel across all possible patch locations.

\paragraph{Results}
For our MNIST model,
  a $6 \times 6$ patch is large enough to successfully
  attack 45\% of the images.
The success rate for $4 \times 4$ patches was 19\%,
  and for $8 \times 8$ patches 80\%.
When an image can be attacked, there are often many possible locations
  where an adversarial patch can be placed:
  for a $6 \times 6$ patch, out of all images where a patch attack is
  possible, there were on average of 41 different positions
  where the patch can be placed.

For our Fashion MNIST model,
  the success rate for patch attacks was as follows:
  $4 \times 4$ patch: 27\% success, $5 \times 5$ patch: 50\% success,
  $6 \times 6$ patch: 60\% success.

These results indicate that, on MNIST, an attacker needs to control
a $6 \times 6$ patch to have close to a 50\% chance of success, while a
$5 \times 5$ patch is large enough for Fashion MNIST,
occupying 5\% and 3\% of the images respectively.

As recent work \cite{chiang2020certified} focuses on $5 \times 5$ patches
 for MNIST and CIFAR-10,
 we use $5 \times 5$ patches for all datasets.


%% file: defense.tex

\section{Our Defense}\seclabel{sec:defense}

The basic idea of the minority reports defense is to occlude part of the
image and classify the resulting image.
If the occlusion completely covers the adversarial patch,
then the attacker will be unable to influence the classifier's prediction.
We don't know where the adversarial patch might be located, so we
stride the occlusion area across the image.
Because we use an occlusion area sufficiently larger than the adversarial patch,
no matter where the adversarial patch is placed
there should be a cluster of occlusion positions that all yield the
same prediction.

\subsection{Creating a prediction grid}
Our defense first generates a \emph{prediction grid},
then analyzes it for patterns that indicate an attack.
We generate the prediction grid as follows.
For defending MNIST images against a $5 \times 5$ adversarial patch,
we use a $7 \times 7$ occlusion region.
We slide the $7 \times 7$ occlusion region over the $28 \times 28$ image
with a stride of one pixel, yielding $22 \times 22$ possible locations
for the occlusion region.
The prediction grid is a $22 \times 22$ array that records,
for each location, the classifier's output.
At each location, we mask out the corresponding occlusion region of the
image, classify the occluded image, obtain the confidence scores from
the classifier's softmax layer, and record that in the corresponding
cell of the prediction grid.
Cell $(i,j)$ of the prediction grid contains the confidence scores
for all 10 classes, when the pixels in the square $(i,j), \dots, (i+7,j+7)$
of the image are masked out.

We visualize the pattern of occlusions in \Figref{fig:into:attack_obscure},
  though with a large stride for illustration.
A stride of one on MNIST
  produces prediction grids such as \figref{fig:intro:grid_examples}
  and \figref{fig:benign_scattered_pred, fig:benign_clustered_pred}.
  
\begin{figure}[tb]\begin{center}
\begin{subfigure}[t]{0.49\linewidth}
  \centering
  \includegraphics[width=\linewidth]{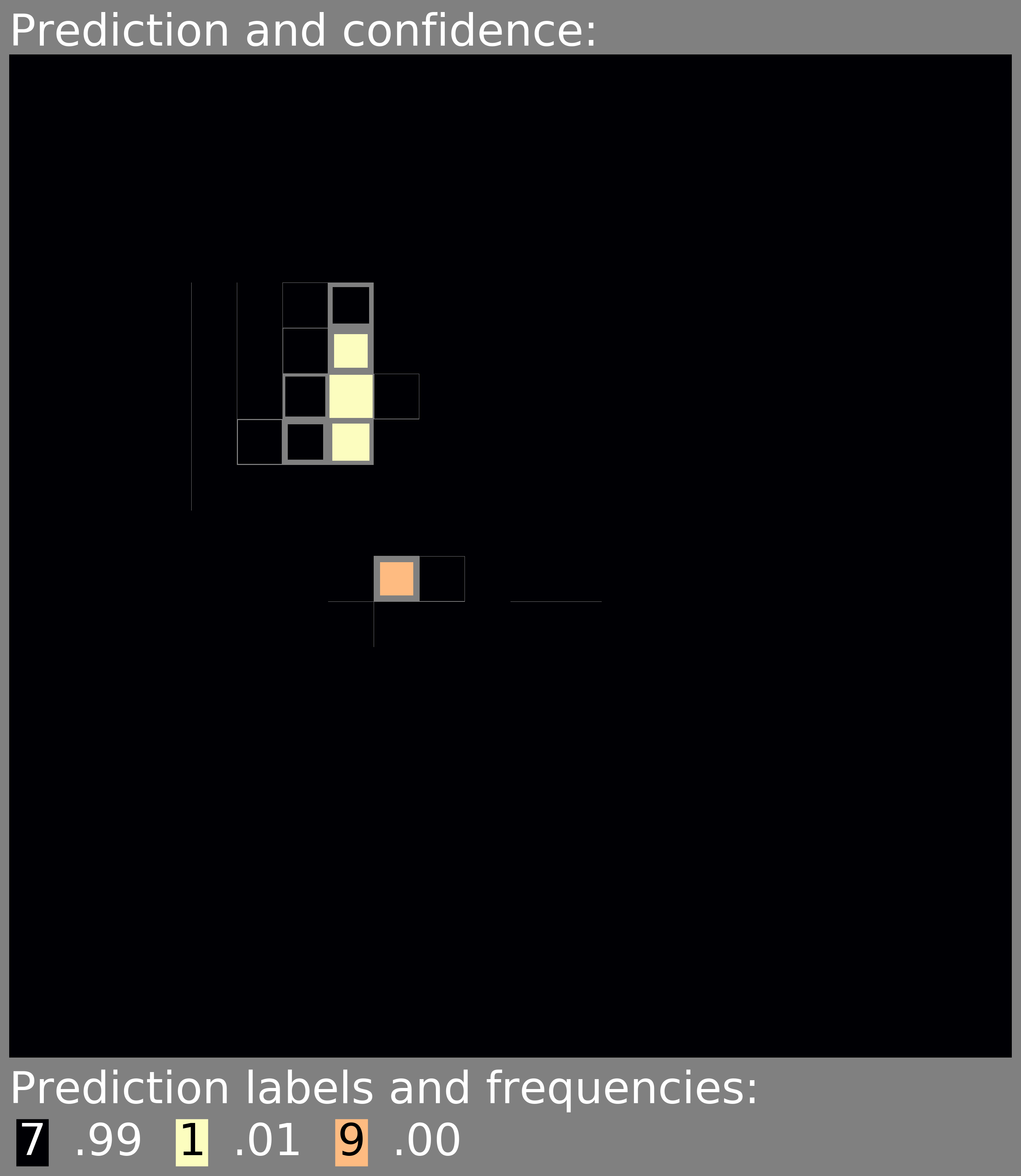}  
  \subcaption{Scattered minority predictions}
  \figlabel{fig:benign_scattered_pred}
\end{subfigure}
\begin{subfigure}[t]{0.49\linewidth}
  \centering
  \includegraphics[width=\linewidth]{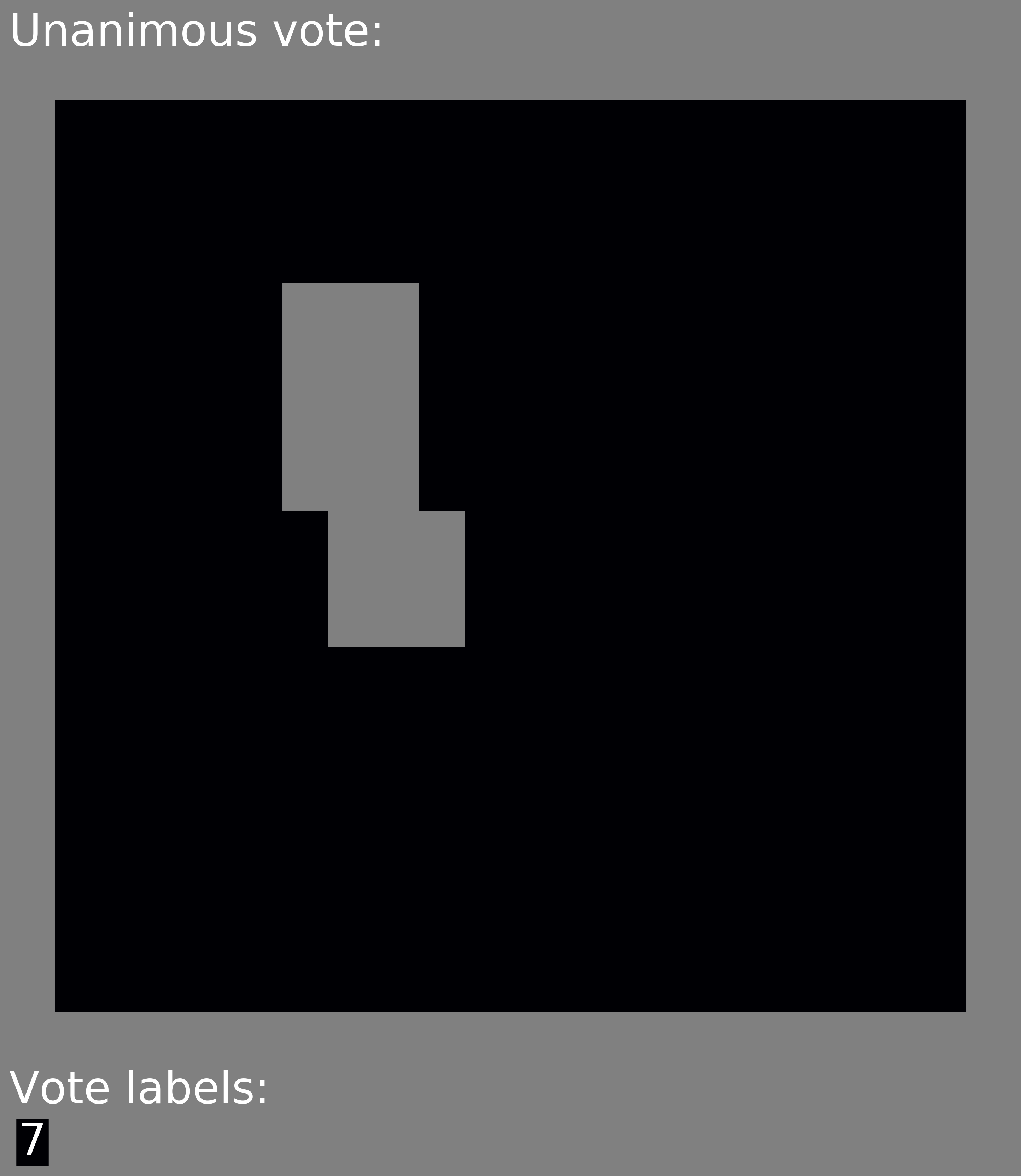}  
  \subcaption{Vote grid: benign}
  \figlabel{fig:benign_scattered_vote}
\end{subfigure}
 
\begin{subfigure}[t]{0.49\linewidth}
  \centering
  \includegraphics[width=\linewidth]{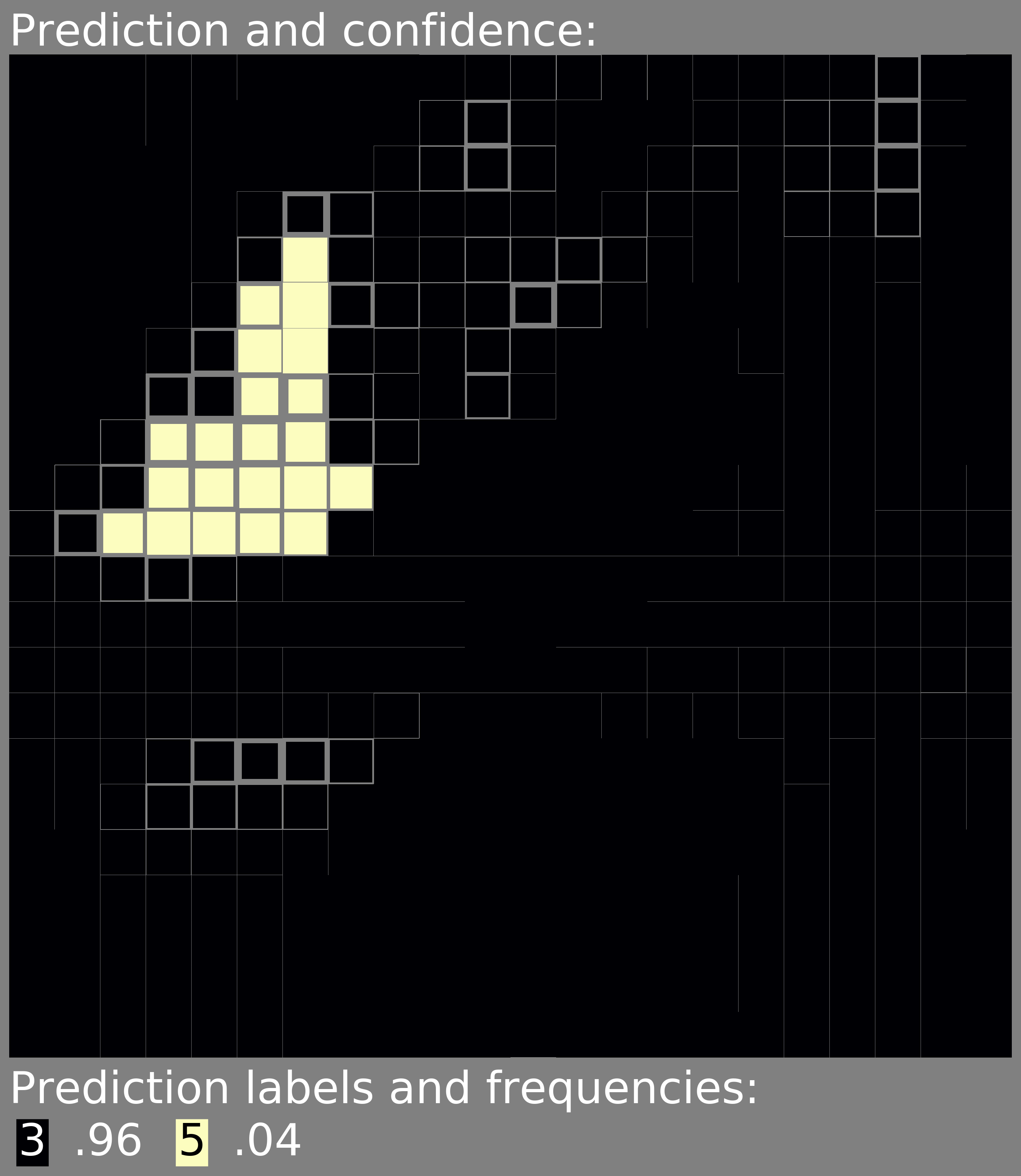}  
  \subcaption{Cluster of minority predictions}
  \figlabel{fig:benign_clustered_pred}
\end{subfigure}
\begin{subfigure}[t]{0.49\linewidth}
  \centering
  \includegraphics[width=\linewidth]{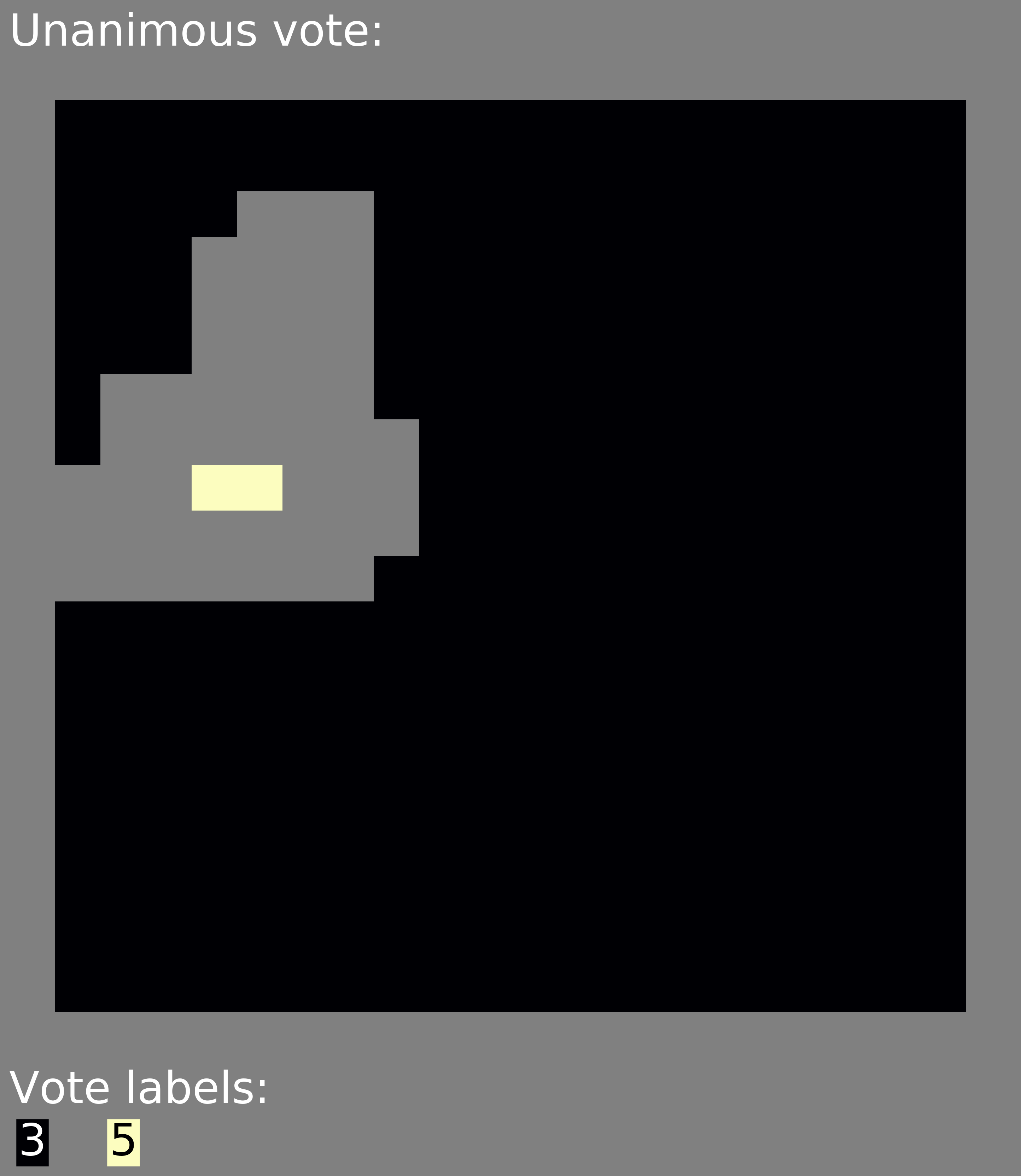}  
  \subcaption{Vote grid: attack}
  \figlabel{fig:benign_clustered_vote}
\end{subfigure}
\caption{In (a) and (c), we show the prediction grids for two benign
images.  (b) and (d) show the corresponding vote grids.
We must decide if the minority votes (yellow) are benign errors 
or what remains of the truth after an attack has influenced the other
predictions.
Unanimous voting classifies the top example as benign and the 
bottom as an attack.
}
\figlabel{fig:prediction_and_voting}
\end{center}\end{figure}

If the image contains an adversarial patch centered at location $(i,j)$,
then obscuring at each of the 9 locations centered at $(i-1,j-1),\dots,(i+1,j+1)$
yields nine images where the adversarial patch has been completely overwritten,
and the predictions in those cells of the prediction grid are completely
unaffected by the attacker.
If the classifier is sufficiently accurate on occluded images,
we can hope that all of those 9 predictions match the true label.
Thus, within the prediction grid, we can expect to see a $3 \times 3$
region where the predictions are uninfluenced by the attacker
and (hopefully) all agree with each other.
Our defense takes advantage of this fact.

\subsection{Detection}
In a benign image, typically every cell in the prediction grid
  predicts for the same label.
In contrast, in a malicious image, we expect there will be a
  $3 \times 3$ region in the prediction grid (where the adversarial
  patch is obscured) that predicts a single label, and some
  or all of the rest of the prediction grid will have a different
  prediction.
We use this to detect attacks.

In our simplest defense, we look at all $3 \times 3$ regions
  in the prediction grid that vote unanimously for the same
  label (i.e., all 9 cells yield the same classification).
If there are two different labels that both have a $3 \times 3$
  unanimous vote, then we raise an alarm and treat this as a
  malicious image.

Equivalently, we categorize each $3 \times 3$ region within the
  prediction grid as either unanimously voting for a class (if all
  9 cells in that region vote for that class) or abstaining
  (if they don't all agree).
We construct a $20 \times 20$ \emph{voting grid} recording these
  votes.
If the voting grid consists of solely a single class and abstentions,
  then we treat the image as benign, and we use that class
  as the final prediction of our scheme.
Otherwise, if the voting grid contains more than one class,
  we treat it as malicious.

The idea behind this defense is twofold.
First, in a benign image, we expect it to be rare for any
  $3 \times 3$ region in the prediction grid to vote unanimously
  for an incorrect class: that would require the classifier to be
  consistently wrong on 9 occluded images.
Therefore, the voting grid for benign images will likely contain
  only the correct class and abstentions.
Second, for a malicious image, no matter where the adversarial
  patch is placed, there will be a $3 \times 3$ region in the prediction
  grid that is uninfluenced by the attack and thus can be expected
  to vote unanimously for the true class.
This means that the voting grid for malicious images will likely contain
  the correct class at least once.
This places the attacker in an impossible bind:
if the attack causes any other class to appear in the voting grid,
  the attack will be detected;
but if it does not, then our scheme will classify the image correctly.
Either way, the defender wins.

We can formulate our defense mathematically as follows.
Let $x$ denote an image, $m_{i,j}$ denote the mask that occludes pixels in
$[i,i+7] \times [j,j+7]$, and $x \odot m_{i,j}$ denote the result of
masking image $x$ with mask $m_{i,j}$.
Then the prediction grid $p$ is constructed as
\begin{equation}
p_{i,j} = C(x \odot m_{i,j}, m_{i,j}),
\end{equation}
where the classifier $C$ outputs a vector of confidence scores.
The voting grid is defined as
\begin{equation}
v_{i,j} =
\begin{cases}
c &\text{if } c = \arg\max_{c'} p_{i+u,j+v,c'} \; \forall  u,v \in \{0, 1, 2\} \\
\textvisiblespace &\text{otherwise.}
\end{cases}
\end{equation}

If there exists a single class $c$ such that $v_{i,j}=c$ or
$v_{i,j}=\textvisiblespace$ for all $i,j$, then our scheme treats the
image as benign and outputs the class $c$; otherwise, our scheme
treats the image as malicious.

We illustrate how the defense works with two examples.
For instance, if the prediction grid is as shown
  in \figref{fig:benign_scattered_pred}
  then it yields the voting grid in 
  \figref{fig:benign_scattered_vote}.
This will be treated as benign, with classification 7.
We show another example of a prediction grid in 
  \figref{fig:benign_clustered_pred}
  and the resulting voting grid in
  \figref{fig:benign_clustered_vote}.
This image will be treated as malicious, and our scheme will
decline to classify it.  In particular, it is possible that the
true label is \texttt{5}, but an adversarial patch was placed
in the upper-left that caused most of the classifications to be
shifted to \texttt{3}, except for a few cases where the patch was
partly or wholly obscured.
It is of course also possible that the image was benign and a cluster
of classification errors caused this pattern, which is the case here.

\subsection{Visualization}
To give some intuition, we visualize a few sample prediction grids in
\figref{fig:prediction_examples}.
The $22 \times 22$ prediction grid is displayed as a Hinton diagram with
$22 \times 22$ squares.
The color of each square indicates which class had highest confidence
at that location in the prediction grid
(i.e., the class predicted by the classifier).
The size of each square is proportional to the confidence of that class.
  
\begin{figure}[htb!]\begin{center}

\begin{subfigure}[t]{0.49\linewidth}
  \centering
  \includegraphics[width=\linewidth]{figs/mnist/MNIST_benign_max2_18}  
  \subcaption{Benign: representative}
  \figlabel{fig:occluded_benign_1}
\end{subfigure}
\begin{subfigure}[t]{0.49\linewidth}
  \centering
  \includegraphics[width=\linewidth]{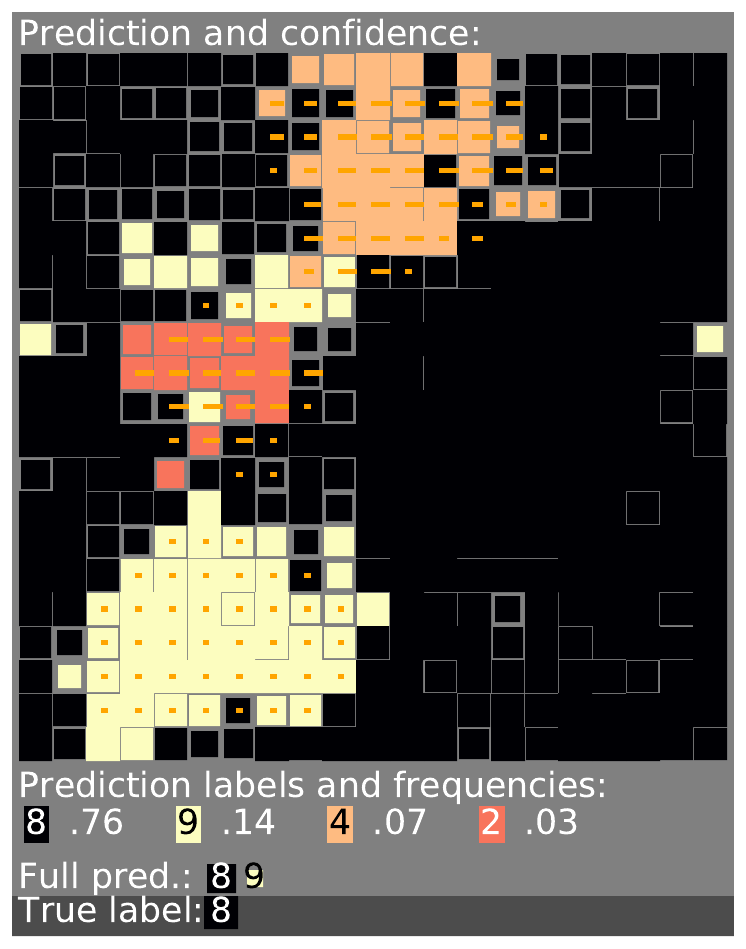}  
  \subcaption{Benign: uncommon}
  \figlabel{fig:occluded_benign_2}
\end{subfigure}
 
\begin{subfigure}[t]{0.49\linewidth}
  \centering
  \includegraphics[width=\linewidth]{figs/mnist/MNIST_attack_max8_38}  
  \subcaption{Attack}
  \figlabel{fig:occluded_attack_1}
\end{subfigure}
\begin{subfigure}[t]{0.49\linewidth}
  \centering
  \includegraphics[width=\linewidth]{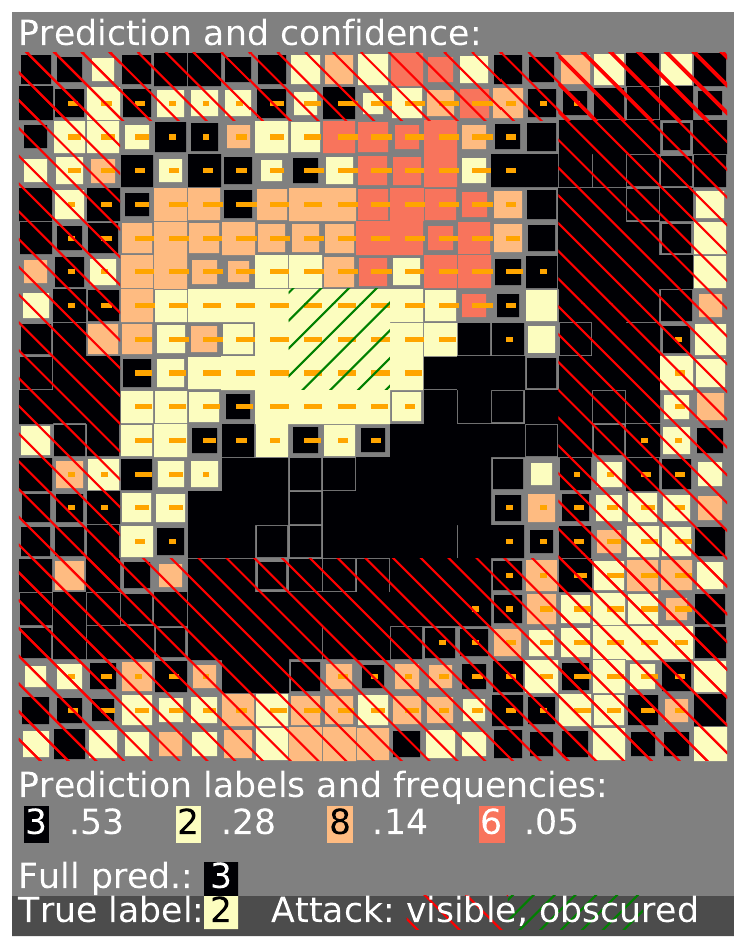}  
  \subcaption{Attack}
  \figlabel{fig:occluded_attack_2}
\end{subfigure}
\caption{
Representative prediction grids for benign and undefended attack MNIST images.
Color indicates the arg max label for that occlusion position and
confidence is indicated by how much of the square is filled.
We show at the bottom of each figure a legend indicating which class
  each color corresponds to and its frequency in the prediction grid;
  we also show the top prediction and confidence if no pixels are occluded.
For attack images, green hashes show the $3 \times 3$ grid of predictions that completely occlude 
the attack, red hashes show the predictions that do not occlude the attack at all.
The hashes are not part of our defense, merely an aid for the reader.
(The short orange bars are from a detection method that compares with the non-occluded prediction.)
}
\figlabel{fig:prediction_examples}
\end{center}\end{figure}

We show a representative example from each of
four different common cases that we have seen:
\begin{enumerate}
\item[(a)] Most benign images have a prediction grid that 
 predicts all for the same label or 
 has just scattered minority predictions and looks like case (a):
  the predictions almost always agree with the true label, for
  almost all positions of the occlusion region, but there are a few
  locations that when occluded cause classification errors (non-black squares).
  These will be correctly classified and treated as benign by our scheme.
\item[(b)] A few benign images have prediction grids that are more noisy
  and contain large clusters of incorrect predictions in the prediction
  grid.
  These will be (incorrectly) categorized as malicious by our scheme, i.e.,
  they will cause a false positive.
\item[(c)] We show the prediction grid resulting from a typical attack
  image, with a adversarial patch placed near the center of the image.
  The green cross-hatching represents the locations that completely occlude
  the adversarial patch.
  Those locations in the prediction grid, as well as some other locations
  in a broader ring around this, vote unanimously for the true label
  (\texttt{1}).
  Occlusion regions placed elsewhere fail to occlude the adversarial
  patch and cause the classifier to mis-classify the image as the
  attacker's target class (\texttt{2}).
  Our scheme correctly recognizes this as malicious, because
  the voting grid contains both unanimous votes for \texttt{1} and for
  \texttt{2}.
\item[(d)] Other attack images have even more noise outside the fully
  occluded area.
  These too are correctly recognized as malicious, because the
  voting grid contains unanimous votes for multiple labels,
   here 3, 2, and 6.
\end{enumerate}

\subsection{The full minority reports defense}
We found that the above defense can be improved by
incorporating two refinements: (a) using soft agreement instead of
hard unanimity, and (b) tolerating outliers.

First, instead of checking whether a $3 \times 3$ region in the
prediction grid votes unanimously for the same label, we
check whether the confidence for that label, averaged over
the region, exceeds some threshold.
For instance, with a $90\%$ threshold, if the confidence scores
for class $c$ within that $3 \times 3$ region average to $0.9$ or larger,
then we'd record a vote for $c$ in the voting grid; if no class
exceeds the threshold, then we record an abstention.

Second, when computing the average, we discard the lowest
score before computing the average.
This allows us to tolerate a single outlier when checking for
agreement in a $3 \times 3$ region.

Mathematically, we fix a threshold $\tau$, and then form the
voting grid as
\begin{equation}
v_{i,j} = \begin{cases}
c &\text{if } \text{avg}(\{p_{i+u,j+v,c} \; \forall  u,v \in \{0, 1, 2\} \})
\ge \tau\\
\textvisiblespace &\text{otherwise.}
\end{cases}
\end{equation}
Here we define $\text{avg}(S)$ to be the average of
$S \setminus \{\min S\}$, i.e., the average of all but the lowest score
in the multiset $S$.

The threshold $\tau$ is a hyper-parameter that can be used to
control the trade-off between false positives and false negatives.
Increasing $\tau$ reduces the number of false positives, but
also risks failing to detect some attacks;
decreasing $\tau$ increases detection power, at the cost of increasing
the false positive rate.

The size of the occlusion region is another hyper-parameter of
our defense.
In our experiments, we always chose an occlusion region that is two
pixels larger than the largest adversarial patch we seek to defend.
Thus our occlusion region will be $7 \times 7$ and
  we provide certified results against adversarial patches up to $5 \times 5$ in size.
Our approach can generalize to other shapes, such as rectangles or
even to arbitrary shapes, so long as they are known in advance.
We can defend against a rectangular $w \times h$ sticker,
with a $(w+2)\times (h+2)$ occlusion region.
To defend against stickers with some other known shape $S$,
the occlusion region can be obtained as the union of 9 translations
of $S$, where we translate independently by $-1$, $0$, or $1$ pixels
in each dimension.

We visualize the operation of our final defense in \figref{fig:soft_vote}.

\begin{figure}[tb]\begin{center} 
\begin{subfigure}[t]{0.49\linewidth}
  \centering
  \includegraphics[width=\linewidth]{figs/mnist/img0_pred}  
  \subcaption{Cluster of minority predictions}
  \figlabel{fig:benign_clustered_pred_soft}
\end{subfigure}
\begin{subfigure}[t]{0.49\linewidth}
  \centering
  \includegraphics[width=\linewidth]{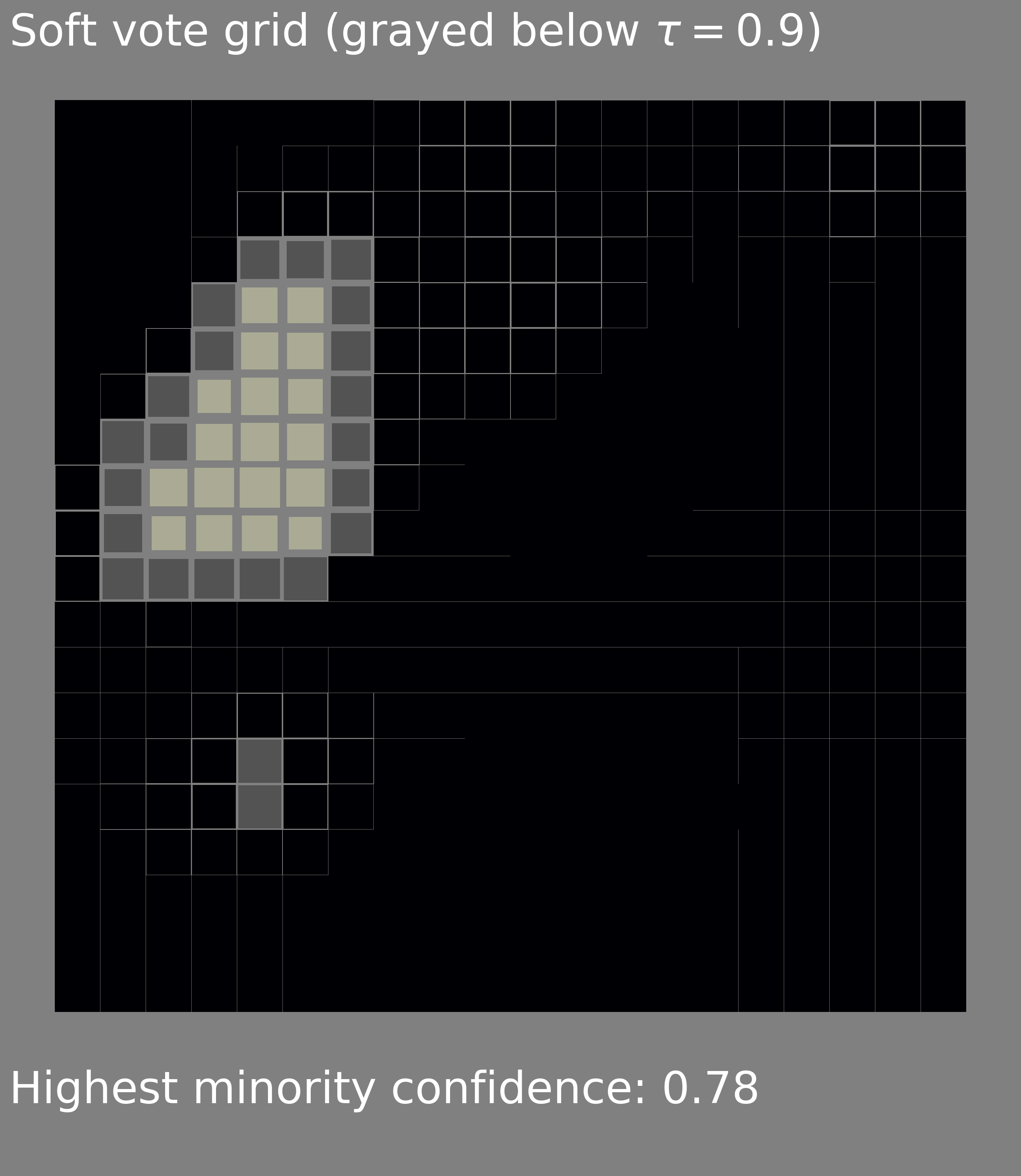}  
  \subcaption{Vote grid: benign}
  \figlabel{fig:benign_clustered_vote_soft}
\end{subfigure}
\caption{
Our full defense on the benign prediction grid from 
  \figref{fig:benign_clustered_pred},
 with $\tau$ = 0.9 classifying as benign (b).
 A sticker under any of the non voting areas would be undetected.
 A sticker in the lower right, when occluded,
 would leave in (a) the confident remains of the original prediction,
 and be classified as an attack.
}
\figlabel{fig:soft_vote}
\end{center}\end{figure}


%% file: security.tex

\section{Security Evaluation}\seclabel{sec:security}

One benefit of our design is that it enables us to
guarantee the security of our scheme on some images.
We describe our certified security analysis in this section.

The core observation is: if the adversarial patch is
completely occluded, then the adversary cannot have any
influence on the prediction made by the classifier on the
corresponding occluded image.
For certified security, we make a very conservative assumption:
we assume that the adversary might be able to completely control
the classifier's prediction for all other occluded images
(i.e., where the patch is only partly occluded, or is not
occluded at all).
This assumption lets us make a worst-case
analysis of whether the classification a particular image
could change in the presence of an adversarial patch of a particular
size.

Notice that wherever the sticker is placed, there
will be a $3 \times 3$ grid in the prediction grid that is
unaffected by the sticker.
(This is because with a stride of one we use an occlusion region that is 2 pixels
larger than the maximum possible sticker size.)
It follows that there will be some cell in the voting grid
that is not changed by the sticker.

If the voting grid for an image $x$ is completely filled
with votes for a single class $c$, with no abstentions, then
any image $x'$ that differs by introduction of a single sticker
will either be classified by our defense as class $c$ or will
be detected by our defense as malicious.
(This follows because at least one element of the voting grid
is unaffected by the sticker, so at least one element of the
voting grid for $x'$ will vote for $c$.
If no other class appears in the voting grid, then our defense
will classify $x'$ as class $c$; if some other class appears,
then our defense will treat $x'$ as malicious.)
Thus, such images can be certified safe---there is no way
to attack them without being detected.
If the prediction is also correct, we classify the image as 
 certified accurate.

In contrast, if the voting grid has even one region that does not vote,
or votes as the attacker would like,
then our conservative analysis is forced to assume that it might
be possible to attack the image:
the attacker can place a sticker at that location,
potentially changing all the other regions' votes, and
thereby escape detection.

We evaluate the security of our scheme by
measuring the fraction of images that can be certified safe
and certified accurate,
according to the conservative analysis above.


%% file: higher_res.tex

\section{Higher Resolution Images}

For higher resolution images, a stride of one pixel becomes prohibitive.
Increasing the stride lets us manage the cost.
For a patch of size $p \times p$ pixels and a stride of $s$ pixels,
  an occlude area of $(p + 2s) \times (p + 2s)$  produces
  nine full occlusions of any patch, if the patch is aligned to our stride grid.
This mirrors what we have done with a stride of one.
To account for patches not aligned to our stride grid, 
  we increase our occlude by one stride.
Thus our occlude area is $(p + 3s) \times (p + 3s)$ pixels for a stride of s, 
   for $s > 1$.

As an example, if CIFAR-10 had twice the resolution, 
  our $5 \times 5$ pixel patch would be $10 \times 10$ pixels.
With a stride of two, our occlude area would be $16 \times 16$, or $(10 + 3 \times 2 ) \times (10 + 3 \times 2)$.
  


%% file: defense_experiments.tex

\section{Experiments}\seclabel{sec:experiments}
We evaluate the effectiveness of our defense by measuring the 
clean accuracy 
  (the images that when unmodified are classified correctly by class and as benign)
and the certified accuracy 
  (the images that when unmodified are classified correctly by class and as benign 
  and where any attack -- targeted or un-targeted --
   will either not change the classification or will be detected).

\paragraph{Method}
We measure the clean and the certified accuracy
  on the 5000 or 6000 validation images.
We perform multiple trials, using a different
  random 90/10 train/validation split for each trial.
 For each dataset we perform $n = 4$ trials.
The standard deviation is relatively low
  (for clean and certified accuracy they are
  CIFAR-10: 0.2 -- 0.8\% 0.5 -- 1.1\%,
  Fashion MNIST:  0.2--0.4\% 0.2--0.6\%,
  MNIST:   0.0 -- 0.1\% 0.1--0.5\%).
We report results for different points in the tradeoff
  between clean and certified accuracy,
  and compare with 
  recent related work using Interval Bounds Propagation (IBP) \cite{chiang2020certified}.

\paragraph{Results}
Our results, \tabref{tab:exp_res},
 show that our defense achieves relatively high clean and certified accuracy
 and outperforms the previous state of the art.
\input{figs/defense_exper_table}

For CIFAR-10, 
 we achieve a clean accuracy of 92.4\% and
  43.8\% of images can be certified accurate
  (no matter where a sticker is placed, the resulting
  image will either be classified correctly or the attack will be
  detected) for $5 \times 5$ stickers.
This is significantly better than recent work by Chiang et al. \cite{chiang2020certified},
  which achieves clean accuracy of 47.8\% and certified accuracy
  of 30.3\% for CIFAR-10 against $5 \times 5$ stickers.

For MNIST,  
  we achieve a clean accuracy of 99.4\% 
  and 64.2\% of images can be certified accurate for $5 \times 5$ stickers.
This is again significantly better than recent work \cite{chiang2020certified}:
  the error rate on clean images is more than an order of magnitude lower,
  and the certified accuracy is slightly higher.

Our measurement of certified accuracy is based
  on conservative assumptions.
We suspect that many images that we cannot certify accurate are in
  fact secure against attack, even though we cannot prove it.
Thus, the number
  certified accurate represents a conservative lower bound on the
  true robustness of our scheme.

\paragraph{Discussion}
Our experiments show that by choosing a high $\tau$, 
  we can achieve clean accuracy that is very close to
  the accuracy of our inner model on non-occluded images.
With a lower $\tau$ we can achieve a higher certified accuracy 
  at the cost of a lower clean accuracy.
  
For CIFAR-10, the architecture we used is reported to have an accuracy of 94.0\%
when trained appropriately.
We did not replicate all aspects of the authors' training procedure, and achieved only 92.5\%.
Once we replicate their full training procedure, we expect our CIFAR-10 results would also improve.

We did an ablation study where we omitted the occlude training, and found that the occlude
training is essential: Without it, the defense is extremely ineffective.


%% file: figs/defense_exper_table.tex
\begin{table}[htp] 
\caption{
  The clean accuracy and certified accuracy of our defense (MR)
  vs the previous state of the art (IBP) on all three datasets,
  for a $5 \times 5$ adversarial patch.
  We report the literature reported accuracy of our inner model architectures in the third column.
  We report the accuracy our inner model achieves on non-occluded clean images
  in the fourth column.
  }
\begin{center}
\begin{tabular}{@{}llrrrr@{}}
\toprule
                              &                                   &  \multicolumn{4}{c}{Accuracy}   \\\cmidrule[.3pt]{3-6}
\multicolumn{2}{l}{Dataset \quad Defense} & \multicolumn{1}{l}{Lit.} & \multicolumn{1}{l}{Inner} &  \multicolumn{1}{l}{Clean} &  \multicolumn{1}{l}{Cert.} \\
 \midrule
 CIFAR-10  &  
                      \multicolumn{2}{l}{IBP \cite{chiang2020certified}}  &                   & 47.8\% & 30.3\% \\\cmidrule[.3pt]{2-6}
                   &   MR (Our) &   94.0\%               & 92.5\% & 78.8\% & 77.6\% \\
                   &                   &                  &             & 90.6\% & 62.1\% \\
                   &                   &                  &               &  \textbf{92.4}\%  &  \textbf{43.8}\% \\
    \midrule[.3pt]
                 
 Fashion & MR           &                  & 93.8\%  &  85.4\% & 84.3\% \\
               &                  &                  &             & 93.0\% & 69.4\% \\
              &                   &                  &             & 93.9\%  &  42.0\% \\
  \midrule[.3pt]
  
 MNIST 
               &  \multicolumn{2}{l}{IBP \cite{chiang2020certified}}     &                  & 92.9\% & 62.0\% \\
  \cmidrule[.3pt]{2-6}
              & MR         &    99.6\%   &  99.6\% & 95.1\%  & 94.9\%\\
             &                 &                  &             & 99.0\%  & 75.8\% \\
             &                 &                  &             & \textbf{99.4}\%  & \textbf{64.2}\% \\

\bottomrule
\end{tabular}
\end{center}
\tablabel{tab:exp_res}
\end{table}%

%% file: accuracy_effects.tex

\section{Effects of Occlude Training}\seclabel{sec:accuracy_effects}

Our defense requires the inner model to handle occluded images well.
To assess the effect of this requirement,
   we trained models with and without occlusions 
   for all three inner-model architectures.

Training on occluded images appears to have only a small change on the accuracy of the inner model on non-occluded images,
  see \tabref{tab:archs_acc}.
The change is at worst the standard deviation of our measurements.
Note from \tabref{tab:exp_res} that the clean accuracy of our defense
  might have either a small or no drop from the accuracy of our inner-model.

\newcolumntype{L}{>{$}l<{$}}

\begin{table}[htp]
\caption{
The effect of training on occluded images,
on the inner model's accuracy on non-occluded images.
We show the difference (last column) and the standard deviation ($n=4$).
  }
\begin{center}
\begin{tabular}{@{}lLLl@{}}
\toprule 
    & \multicolumn{2}{c}{Type of training images}  & 
 \\
Dataset  & \text{Non-occluded} & \text{Occluded} &  $\Delta$ 
 \\\midrule
CIFAR-10          & 92.5\pm0.3\% & 92.5\pm0.2\%    & $-0.0\%$ \\ 
Fashion             & 94.1\pm0.4\% & 93.8\pm0.3\%   & $-0.3\%$\\  
MNIST               & 99.58\pm0.08\% & 99.63\pm0.33\%  & $+0.05\%$ \\ 
\bottomrule
\end{tabular}
\end{center}
\tablabel{tab:archs_acc}
\end{table}%

Note that this does not measure the accuracy of our defense as a whole.
Our defense feeds the inner model occluded images at test time,
and accuracy on occluded images is slightly lower than on non-occluded images.


%% file: related.tex

\section{Related Work}

In earlier work, Hayes proposes a defense against sticker attacks
using inpainting of a suspected sticker region to remove the sticker
from the image \cite{Hayes_2018_DigWatermark}.
This is similar to our defense.
However, Hayes uses a heuristic to identify the region to inpaint
  (based on unusually dense regions within the saliency map),
  so any attack that fools the heuristic could defeat their defense.
One could use inpainting in our scheme instead of occlusion, and
  it is possible this might improve accuracy, though our work
  can be viewed as showing that simple occlusion suffices to get
  strong results.
Naseer et al. propose a defense against sticker attacks by smoothing
  high frequency image details to remove the sticker
  \cite{Naseer-etal-LGS-2018}.
They limit accuracy loss by using windows that overlap by a third,
  but their windows are smaller than the attack patch.
Chiang et al. broke both of these defenses \cite{chiang2020certified},
  so neither is effective against adaptive attacks;
  in contrast, we guarantee security against adaptive attack.

In concurrent work,
Wu et al.   
 defend against adversarial patches with adversarial training
\cite{wu2020defending}.
The primary advantage of our approach is that it provides certified
  security.

In concurrent work,
  Chiang et al. study certified security against patch attacks
  using interval bounds propagation \cite{chiang2020certified}.
As discussed above, our defense achieves significantly better certified
  accuracy on both MNIST and CIFAR than their scheme.
They also examine how their defense generalizes to other shapes of
  stickers and how to achieve security against $L_0$-bounded attacks,
  topics that we have not examined.


%% file: conclusion.tex


\section{Conclusion}

We propose the minority reports defense, 
  a network architecture designed specially to be
  robust against patch attacks.
We show experimentally that it is successful at defending
  against these attacks for a significant fraction of images.
  
